\documentclass[10pt,journal,compsoc]{IEEEtran}

\ifCLASSOPTIONcompsoc
  \usepackage[nocompress]{cite}
\else
  \usepackage{cite}
\fi

\ifCLASSINFOpdf

\else

\fi

\usepackage{graphicx}
\usepackage{epstopdf}
\usepackage{booktabs}
\usepackage{subfigure}
\usepackage{amsfonts}
\usepackage{amsmath}
\usepackage{algorithm, algorithmic}
\usepackage{subfigure}
\usepackage{cite}

\usepackage{amsthm}
\usepackage[T1]{fontenc}
\usepackage[utf8]{inputenc}
\usepackage{authblk}
\usepackage{float}
\usepackage{hyperref}
\usepackage{ragged2e}
\usepackage{booktabs,makecell, multirow, tabularx}
\hypersetup{hypertex=true,
            colorlinks=true,
            linkcolor=blue,
            anchorcolor=blue,
            citecolor=blue}
\begin{document}
\title{An Evolution Kernel Method for Graph Classification through Heat Diffusion Dynamics}

\author[1]{Xue Liu}
\author[2]{Dan Sun}
\author[2, 1, 3, 4, *]{Wei Wei \thanks{Corresponding author: weiw@buaa.edu.cn}}
\author[1, 3, 4]{Zhiming Zheng}

\affil[1]{Institute of Artificial Intelligence, Beihang University, Beijing, 100191, P. R. China}
\affil[2]{School of Mathematical Sciences, Beihang University, Beijing, 100191, P. R. China}
\affil[3]{Key Laboratory of Mathematics, Informatics and Behavioral Semantics, Ministry of Education, 100191, P. R. China}
\affil[4]{Zhongguancun Laboratory, Beijing, 100094, P. R. China}

\markboth{Preprint submitted to Journal of IEEE TRANSACTIONS ON NETWORK SCIENCE AND ENGINEERING, 2023}%
{Shell \MakeLowercase{\textit{et al.}}: Bare Demo of IEEEtran.cls for Computer Society Journals}

\IEEEtitleabstractindextext{%
\begin{abstract}
Autonomous individuals establish a structural complex system through pairwise connections and interactions. Notably, the evolution reflects the dynamic nature of each complex system since it recodes a series of temporal changes from the past, the present into the future. Different systems follow distinct evolutionary trajectories, which can serve as distinguishing traits for system classification. However, modeling a complex system's evolution is challenging for the graph model because the graph is typically a snapshot of the static status of a system, and thereby hard to manifest the long-term evolutionary traits of a system entirely. To address this challenge, we suggest utilizing a heat-driven method to generate temporal graph augmentation. This approach incorporates the physics-based heat kernel and DropNode technique to transform each static graph into a sequence of temporal ones. This approach effectively describes the evolutional behaviours of the system, including the retention or disappearance of elements at each time point based on the distributed heat on each node. Additionally, we propose a dynamic time-wrapping distance GDTW to quantitatively measure the distance between pairwise evolutionary systems through optimal matching. The resulting approach, called the Evolution Kernel method, has been successfully applied to classification problems in real-world structural graph datasets. The results yield significant improvements in supervised classification accuracy over a series of baseline methods.



\end{abstract}

\begin{IEEEkeywords}
Complex network, Graph spectral theory, Heat kernel, Kernel method, Graph classification
\end{IEEEkeywords}}

\maketitle

\IEEEraisesectionheading{\section{Introduction}\label{sec:introduction}}
\IEEEPARstart{I}{n} real world, many complex systems are comprised by a multitude of autonomous individuals and their interactions. For examples, the animal population networks~\cite{network-social-animal, network-Social-Dolphin} attempt to characterize the social framework through individuals' observable behaviors like association and aggression; the opinion formation social networks~\cite{network-social-opinion} provide insight into the social mechanisms that rule opinions exchanging actions among individuals; and the World Wide Web (WWW)~\cite{network-WWW} is constructed by a complex system of numerous hyperlinks connecting Web pages. In many applications, these complex systems are formulated by the graph models, which use nodes to depict individual entities and constructs links to signify the connections between them.

For each complex system, evolution is an intrinsic nature as the interactions between autonomous individuals happen everywhere and all the time. With individuals interacting across the time axis, some underlying but meaningful traits could be fully displayed from statistical or sociological perspectives. For example, the dolphin social network, as an animal population model, evolves with individuals that no longer keep connections with others within the system lapsing into degeneration. In the opinion formation social network model, each individual within the network is initialized with random opinions, which could be changed under the influence of other individuals, and it evolves by opinions changing until an equilibrium state characterized by the existence of one or several opinion groups is reached. For WWW, since instantaneous mass access to a page inevitably brings congestion to the website, a strategy like PageRank~\cite{pagerank} utilizes the variations in link-based neighborhood probability distributions to represent the network evolution.

\begin{figure*}[h]
\centering
\includegraphics[height=6.5cm,width=18cm]{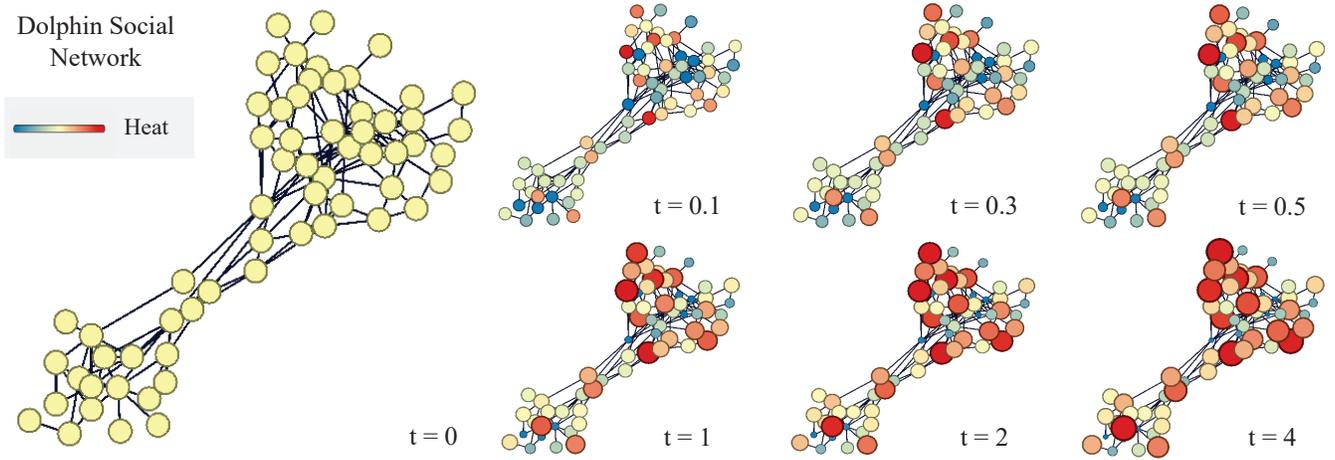}
\caption{The heat map of the dolphin social network evolution displays how the heat changes over time on nodes, with variations in node color and size. Nodes that appear red and larger contain more heat. In the beginning, each node is assigned the same initial heat $u_{0}$ when $t = 0$, then derived by heat kernel $H_{t}$, the network shows different evolutionary features as time $t$ growing from $0.1$, to $4$. Notably, the nodes in the central position absorb more heat from the global scale, resulting in the shrinkage of leaf nodes.}
\label{Figure-Dolphin-Heat-Sequence}
\end{figure*}

How to understand and analyze complex system's evolutional dynamics has attracted much attention in the area of machine intelligence. This is interesting because evolution records a series of temporal snapshots of the varying system from the past, the present into the future. Different evolutional processes result in various dynamical patterns. Intuitively, these evolutionary patterns could act on assessing system's heterogeneities when compared with others. However, the popular graph model has an inherent challenge in manifesting system evolution. Since the graph model is not time-dependent, each graph primarily represents an instantaneous snapshot that depicts the static state of a given system. This limitation makes the static graph unsuitable to demonstrate the long-term evolutionary fashions of the targeted system.

To address this matter, a practical and effective strategy is to implement particular physical mechanisms to convert each static graph into a temporal sequence. For complex systems, the evolution takes place through various interactions between individuals across edges, and this procedure is implemented by the Laplacian matrix transformation in algebra and graph theory. With this mechanism, in this work we opt for the physics-based heat diffusion method to manifest the system evolution. The heat diffusion method is a well-established dynamic model used to represent information propagation on the basis of differential equations and graph spectral theory. This model refers to a dynamic equation $\frac{\partial}{\partial t} u = \Delta u$ to simulate heat or energy propagation processing across the network's edges from the starting node, where $\Delta$ denotes the Laplacian operator. The mechanism behind the model is that the rate of heat propagation is proportional to the volume of heat contained in this area. Heat is constantly exchanged within the system through a diffusion procedure until a permanent state is finally reached, such that for each node, the volume of absorbed heat equals the volume of outflow. Specifically, we consider heat propagation as the essential mean for simulating the information propagation between individuals within the system in this paper.

The dolphin social network in Figure \ref{Figure-Dolphin-Heat-Sequence} offers an insightful example of how heat propagation can illustrate the system evolution. In dolphin society, not everyone has an equal impact on the network. The adult females locate in the central hubs, where information converging, exchanging, and propagation among the population is more efficient. At the same time, the loosely-connected individuals generally situate on the periphery of society. Over time, the characteristics of individuals, such as the heat distribution, gradually become more prominent along the evolutionary trajectory. Adult female members occupying the central parts of society appear to attract males into their community, resulting in a significant proportion of absorbed heat. In contrast, the leaf parts inevitably degenerate due to their negligible impact on the overall structure. Theses individuals present themselves as the gradual dissipation of heat.

This paper examines the graph classification problems, specifically emphasizing the dynamic evolutionary trajectory as a source of heterogeneity when comparing pairwise graphs. In this paper, we proposed the  heat-driven temporal graph augmentation method to convert static graphs into their dynamic sequences and measured the distance between any two sequences. To achieve this, we first introduced heat distribution probability to rank the relative importance of each point in a graph or system at a given moment. Then we utilized the DropNode action to simulate the retention or disappearance of individuals in a system based on the probability weight of each point in the graph. In this way, each static graph could be transferred into a dynamically evolutionary sequence within a predetermined time length. Through the optimal matching of data sequences, we further proposed a time-warping distance GDTW to measure the distance between pairwise dynamically evolving graphs and then constructed a kernel-based classifier to achieve a better separation of the compared graphs ultimately. We conducted a series of real-world supervised classification experiments to validate the effectiveness with 10-fold stratified cross-validation. The experimental results showed our method could successfully strengthen advances in accurate prediction on all datasets.

The key contributions of this publication are multifaceted.
\begin{itemize}
\item[(1)] The heat kernel as the fundamental mechanism for simulating system evolution presents a novel viewpoint for understanding systems' variations and a particular analysis mode for graph classification.
\item[(2)] This paper proposes a time-warping metric to measure the distance between temporal graph episodes so that the graphs with different labels separate effectively.
\item[(3)] Extensive experiments are conducted on real-world datasets to show the effectiveness of supervised classification. Our method could derive a highly accurate, efficient performance on all datasets. Specifically, the classification accuracies can be improved by considering graphs' evolutionary behaviors, rather than relying solely on their static forms.
\end{itemize}

The paper is structured into four main sections. The fundamental definitions of the spectral graph, temporal graph episode, and graph kernel are outlined in Section~\ref{Section-Problem-BACKGROUND-AND-PROBLEM-STATEMENT}. As our focus is on the heat kernel as the evolutionary driver for graphs, thus Section~\ref{Section-Heat-Kernel} delves into the concepts and related properties of the classic heat kernel. In Section~\ref{Section-Evolution-Kernel}, we define the heat-driven temporal graph augmentation and introduce the graph dynamic time warping distance to measure the distance between two graph episodes. Furthermore, we associate the graph dynamic time warping distance with the graph kernel to identify the discrepancies in evolutionary behaviors between two targeted graphs. Finally, Section~\ref{Section-Experiments} provides detailed information and results of comparative experiments conducted on supervised classification.

\newtheorem{definition}{Definition}
\newtheorem{theorem}{Theorem}
\newtheorem{lemma}{Lemma}
\section{BACKGROUND AND PROBLEM STATEMENT}\label{Section-Problem-BACKGROUND-AND-PROBLEM-STATEMENT}
\subsection{Spectral Graph}
We first review basic definitions of complex network graph. Let $G = (V, E)$ denote a network with $n$ nodes in set $V$ and $m$ edges in set $E \subseteq V\times V$, the adjacency matrix $A \in \mathbb{R}^{n \times n}$ encodes the node-wise connection of the network and is defined as follows:
\begin{equation}
     A_{ij}=\left\{
     \begin{aligned}
      1                 &,  \ if \ (v_{i}, v_{j}) \in E, \\
      0                 &,  \ otherwise.
      \end{aligned}
      \right.
\end{equation}

Let $d_{i}$ denote the degree of the vertex $v_{i}$, which has $d_{i}=\sum\nolimits_{j=0}^{n-1}A_{ij}$. We denote the degree matrix of graph $G$ as $D = diag(d_0, d_1, ..., d_{n-1})$, whose diagonal entry having value $d_{i}$, $i = 0 , ..., n-1$ is the degree of node $v_{i}$. The volume of the graph $G$, is given by
\begin{equation}
     \textrm{vol} G = \sum\limits_{i = 0}^{n-1} d_{i}.
\end{equation}

To begin, we consider the Laplacian matrix of graph $G$ which could be defined as $L = D - A$, whose each entry is as follows:
\begin{equation}
     L(i, j) =\left\{
     \begin{aligned}
      d_{i}             &,  \ if \ i=j \\
      -1                &,  \ if \ i\neq j, (v_{i}, v_{j})\in E\\
      0                 &,  \ otherwise
      \end{aligned}.
      \right.
\end{equation}
And the normalized Laplacian matrix $\mathcal{L}$ is defined by the degree matrix $D$:
\begin{equation}
     \mathcal{L} = D^{-\frac{1}{2}} L D^{-\frac{1}{2}} = I - D^{-\frac{1}{2}}AD^{-\frac{1}{2}},
\end{equation}
with the convention $D^{-\frac{1}{2}}(i,i) = 0$ for $d_{v_{i}} = 0$, where $I$ is the identity matrix.

Then the entry of $\mathcal{L}$ can be written as
\begin{equation}
     \mathcal{L}(i, j) =\left\{
     \begin{aligned}
      1                             &,  \ if \ i=j,  d_{i} \neq 0\\
      -\frac{1}{\sqrt{d_{i}d_{j}}}  &,  \ if \ i\neq j, (v_{i}, v_{j})\in E\\
      0                             &,  \ otherwise
      \end{aligned}.
      \right.
\end{equation}

Note that Laplacian matrix $\mathcal{L}$ will be symmetrical when network $G$ is undirected, otherwise it will be asymmetrical, thus it can be factorized as $\mathcal{L} = \Phi \Lambda \Phi^{\mathsf{T}}$, where $\Lambda$ is a diagonal matrix on the sorted eigenvalues $0 = \lambda_{0} \leq \ldots \leq \lambda_{n-1}$ of the corresponding eigenvectors $\phi_{0},\phi_{1},\ldots, \phi_{n-1}$, which are orthogonal to each other and stacked in matrix $\Phi$. The set of eigenvalues $\lambda_{0}, \ldots , \lambda_{n-1}$ is usually called the \emph{spectrum} of $\mathcal{L}$ or the \emph{spectrum} of the associated graph $G$.

\subsection{Temporal Graph Episode}
Evolutionary graphs can be described as a series of transformations of the graph space into itself. The consideration of the system's evolution varies depending on the problem at hand. Edge-centric, vertex-centric, and graph-centric perspectives are used to analyze the system evolution from the standpoint of a specific relation (edge), entity (node), or the global system (the entire graph), respectively.

We consider the graph space $\mathcal{G}$ and a one-parameter mapping $\varphi: \mathbb{R} \times \mathcal{G} \rightarrow \mathcal{G}$, which maps the graph space $\mathcal{G}$ with any time $t \in \mathbb{R}$ into itself satisfying the following constrains:

\begin{itemize}
\item[(1)] $\varphi(t_{0}, G) = G$, where $ t_{0}\in \mathbb{R}$ denotes the initial time, $\forall G \in \mathcal{G}$.
\item[(2)] $\varphi(t, G)$ is continuous, $\forall t \in \mathbb{R}$, $\forall G \in \mathcal{G}$.
\item[(3)] $\varphi(t_{2}, \varphi(t_{1}, G)) = \varphi(t_{1} + t_{2}, G)$, $\forall t_{1},  t_{2} \in \mathbb{R}$, $\forall G \in \mathcal{G}$.
\end{itemize}
In this paper, specifically, we denote the evolutionary graph of $G$ at time $t$ with respect to all $t\in \mathbb{R}$ and $G \in \mathcal{G}$ as
\begin{equation}
     G^{t} := \varphi(t, G).
\end{equation}
Correspondingly, considering a time sequence $T = \{t_{0}, t_{1}, \ldots, t_{N}\}$, the \textbf{temporal graph episode} of the evolutionary $G$ is given by
\begin{equation}\label{random-graph-episode}
  S(G) = \{G^{t_{0}}, G^{t_{1}}, \ldots, G^{t_{N}}\}.
\end{equation}

\subsection{Graph Kernel}
Kernels belong to the category of distance-based functions, aiming to assess the equivalence or dissimilarity between specific items. Suppose $kernel: \mathcal{X} \times \mathcal{X} \longrightarrow \mathbb{R}$ denotes a function associated with a Hilbert space $\mathcal{H}$, satisfying $kernel(x, y) = \langle \varphi(x), \varphi(y) \rangle_{\mathcal{H}}$ for a map $\varphi: \mathcal{X} \longrightarrow \mathcal{H}$. Thus, $\mathcal{H}$ is a reproducing kernel Hilbert space, and $kernel$ is said to be a positively defined kernel. Support Vector Machines (SVM) is a well-known kernel trick in machine intelligence.

The approach for defining kernels on graphs typically utilizes the R-Convolution framework. For any two $G_{i}, G_{j}$ in graph space $\mathcal{G}$, the main idea of graph kernel is to decompose each graph as a collection of substructures and then define a kernel value $kernel(G_{i}, G_{j})$ as a similarity measure on the substructure level. The Weisfeiler-Lehman kernel~\cite{WL-kernel} maps graph data into a Weisfeiler-Lehman sequence, and the node attributes of this sequence represent graph information about topology and labeling. The Weisfeiler-Lehman kernel is wildly employed in graph isomorphism tests on graphs as its runtime scales linearly in the number of edges of the graphs and the length of the Weisfeiler-Lehman graph sequence. The Graphlet kernel~\cite{GK-graphlet-kernel} measures graph similarity by counting common $k$-node graphlets, ensuring the computation complexity is restricted in polynomial time. The Deep Weisfeiler-Lehman and Deep Graphlet~\cite{Deep-graph-kernels} leverage language modeling and deep learning to learn latent representations of sub-structures for graphs. The Persistence Fisher kernel~\cite{Persistence-Fisher} relies on Fisher information geometry to explore persistence diagrams on structural graph pattern recognition. The WKPI~\cite{WKPI} incorporates a weighted kernel for persistence images, together with a metric-learning method to learn the best classifier function for labeled data.

\subsection{Problem Statement}
In this paper, we are researching the way of obtaining temporal graph episode of a static graph through its evolutionary dynamics, and seek how to define the graph kernel between pairwise temporal graph episodes $S(G_{i})$ and $S(G_{j})$, i.e.,
\begin{equation}\label{random-graph-episode}
  kernel(S(G_{i}), S(G_{j})).
\end{equation}

\section{Heat Kernel}\label{Section-Heat-Kernel}

The heat equation, referred to the diffusion equation~\cite{heat-diffusion, heat-kernel-community-detection, Heat-Kernel-diffusionrank}, is commonly used to illustrate the temporal evolution of the density of some quantity, such as heat. In physics, Fourier's law of heat conductivity utilizes heat equation to describe the diffusion of heat and other substances through continuous media. Generally, heat flows from regions of high temperature to those of lower temperature positions. Heat diffusion is widely used to represent information propagation in the information science area, such as the graph spectral theory~\cite{spectral-graph-theory} and the Graph Convolutional Networks~\cite{GCN}.

To begin with, we provide the continuous heat kernel as follows, and then introduce some properties of heat kernels, including the estimation of the upper bounder for the heat equation solution and the approximation of the heat kernel.

\begin{definition}
Let $h(t, x)$ denote the heat kernel of a $d$-dimension Riemannian manifold $M$, and let $u = u(t, x)$ at time $t$ and point $x \in M$ satisfy the heat equation
\begin{equation}
  \frac{\partial}{\partial t} u = \Delta u
\end{equation}
with the Neumann boundary condition $\frac{\partial}{\partial \upsilon} u(t, x) = 0$ for any boundary point $x$, where $\Delta$ denotes the Laplacian operator (or, equivalently, the Laplace-Beltrami operator) in local coordinates $x_{1}, \ldots, x_{d}$, i.e.,
\begin{equation}
  \Delta  = \frac{\partial^{2}}{\partial x_{1}^{2}} + \frac{\partial^{2}}{\partial x_{2}^{2}} + \ldots + \frac{\partial^{2}}{\partial x_{d}^{2}}.
\end{equation}.
\end{definition}

The heat kernel or the fundamental solution of heat equation has an explicit expression
\begin{equation}
  h(t, x) = \frac{1}{(4 \pi t)^{\frac{d}{2}}}e^{-\frac{\|x\|^{2}}{4t}},
\end{equation}
which is also called the Gauss-Weierstrass function~\cite{heat-kernel, diffusion-kernels, sun2009concise-heat-diffusion}.

The definition of the discrete heat kernel for graphs is exactly the parrel of the continuous heat kernel on Riemannian manifolds.
\begin{definition}
The heat kernel is a fundamental solution of the following \emph{heat equation}:
\begin{equation}\label{heat-equation}
     \frac{\partial u_{t}}{\partial t} = - \mathcal{L} u_{t},
\end{equation}
where $u_{t}\in \mathbb{R}^{n}$ represents the heat of each node in $G$ at time $t$, $\mathcal{L}$ denotes the normalized Laplacian matrix.
\end{definition}

Here the Laplacian is associated with the rate of dissipation of heat. For any $t \geq 0$, the \emph{heat kernel} $H_{t}$ of $G$ is defined to be the $n \times n$ matrix
\begin{equation}
     H_{t} = e^{-t \mathcal{L}}.
\end{equation}

When the initial heat $u_{0}$ is initialized at time $t_{0} = 0$, the solution to the heat equation provides the heat at each node at time $t$ with the form
\begin{equation}
     u_{t} = e^{-t \mathcal{L}}u_{0} = H_{t}u_{0},
\end{equation}
where $u_{t}(i, j)$ represents the amount of heat transferred from node $v_{i}$ to $v_{j}$ at time $t$.

\subsection{Properties of Heat Kernel}
Since we could factorize $\mathcal{L}$ as $\mathcal{L} = \Phi \Lambda \Phi^{\mathsf{T}}$, thus the closed-form solution is given by
\begin{equation}
     H_{t} = e^{-t \mathcal{L}} = \Phi e^{-t \Lambda} \Phi^{\mathsf{T}} = \sum \limits_{i = 0}^{n-1}e^{-t\lambda_{i}}\phi_{i}\phi_{i}^{\mathsf{T}},
\end{equation}

\begin{lemma}
  For all node $v_{i}$, $v_{j} \in G$, $i \neq j$, we have
    \begin{equation}
     H_{t}(v_{i}, v_{j}) \geq 0.
    \end{equation}
\end{lemma}

\begin{proof}
  The matrix $\mathcal{A} = I - \mathcal{L}$ has all entries non-negative, thus $e^{t\mathcal{A}}$ has all non-negative entries. Since
\begin{equation}
    H_{t} = e^{-t\mathcal{L}} = e^{-t(I-\mathcal{A})} = e^{-t}e^{t\mathcal{A}},
\end{equation}
all entries of $H_{t}$ are non-negative.
\end{proof}

\begin{lemma}\label{Lemma-3}~\cite{matrix_sensitivity}
Let $M(t)$ be a monotone increasing function defined on $[0, +\infty)$, $A$ and $E$ denote two matrices, and $\|e^{At}\| \leq M(t) e^{\beta t}$ for all $t \geq 0$, then
\begin{equation}
  \begin{aligned}
  \phi(t)   & = \frac{\|e^{(A+E)t} - e^{At}\|}{\|e^{At}\|} \\
            & \leq t \|E\| M^{2}(t) e^{(\beta - \alpha(A) + \|E\| M(t))t}.
  \end{aligned}
\end{equation}
\end{lemma}

Lemma~\ref{Lemma-3} investigates the upper bounder of $\phi(t)$. In the next theorem, we use Lemma~\ref{Lemma-3} to show the estimated heat kernel based on matrix exponential is stable against noise.

\begin{theorem}\label{Theorem-3}
  Assume that $\mathcal{\tilde{L}}$ is a matrix with perturbation for normalized Laplacian matrix $\mathcal{L}$, such that $\mathcal{\tilde{L}} = \mathcal{L} + F$, where $\|F\| \leq \epsilon$. Then if $H_{t}$ and $\tilde{H}_{t}$ are the induced heat kernels from $\mathcal{L}$ and $\mathcal{\tilde{L}}$, respectively, we have
  \begin{equation}
    \|H_{t} - \tilde{H}_{t}\| \rightarrow 0.
  \end{equation}
  Here $\| \cdot\|$ is any matrix norm induced from vector norm or its equivalence.
\end{theorem}

\begin{proof}
    We apply Lemma \ref{Lemma-3} with $M(t) = 1$ and $\beta = \|A\|$, then derive
    \begin{equation}
    \begin{aligned}
    \phi(t) & = \frac{\|e^{(A+E)t} - e^{At}\|}{\|e^{At}\|} \\
            & \leq t \|E\| e^{(\|A\| - \alpha(A) + \|E\|)t}.
    \end{aligned}
    \end{equation}

    The heat kernels $H_{t}$ and $\tilde{H}_{t}$ derived by $\mathcal{L}$ and $\mathcal{\tilde{L}}$ are matrix exponential functions, i.e., $H_{t} = e^{-t \mathcal{L}}$ and $\tilde{H}_{t} = e^{-t \mathcal{\tilde{L}}}$, respectively. By applying $\|F\| \leq \epsilon$, we yield
    \begin{equation}
    \begin{aligned}
    \frac{\|H_{t} -  \tilde{H}_{t}\|}{\|H_{t}\|} & = \frac{\|e^{-t \mathcal{L}} -  e^{-t \mathcal{\tilde{L}}}\|}{\|e^{-t \mathcal{L}}\|} \\
    & \leq t \|\mathcal{L} -  \mathcal{\tilde{L}}\| e^{-t \|\mathcal{L} -  \mathcal{\tilde{L}}\|} \\
    & \leq t \|F\| e^{-t \|F\|} \rightarrow 0.
    \end{aligned}
    \end{equation}
    Then it is easy to derive the conclusion
    \begin{equation}
    \|H_{t} - \tilde{H}_{t}\| \rightarrow 0.
    \end{equation}
\end{proof}

\subsection{Approximation of the Heat Kernel}
A natural way for dealing with complex matrix functions is to approximate the function with its Taylor expansion.
\begin{equation}
\begin{aligned}
     H_{t} & = e^{-t \mathcal{L}} \\
            & = I - t\mathcal{L} + \frac{1}{2!}{t\mathcal{L}}^2 - \ldots \\
            & = \sum\limits_{k = 0}^{\infty}\frac{(-1)^{k}(t \mathcal{L})^k}{k!}.
\end{aligned}
\end{equation}

For small time point $t$ approaching 0, i.e., $t \rightarrow 0$, we could apply the low-order truncated Taylor expansion,
\begin{equation}
\begin{aligned}
     H_{t} & = e^{-t \mathcal{L}} \\
            & = I - t\mathcal{L} + \frac{1}{2!}{t\mathcal{L}}^2 + o(t^{3})\\
            & \approx I - t\mathcal{L} + \frac{1}{2!}{t\mathcal{L}}^2.
\end{aligned}
\end{equation}
While for large time point $t$, it turns to be unreliable to approximate $H_{t}$ using truncated Taylor expansion. Thus we take such strategy to approximate heat kernel by
\begin{equation}
     H_{t} \approx I - e^{-\lambda_{1}t}\phi_{1}\phi_{1}^{\mathsf{T}},
\end{equation}
where $\lambda_{1}$ denotes the second smallest eigenvalue, which is also called the \emph{Fiedler value}, and corresponding eigenvector $\phi_{1}$ the \emph{Fielder vector}~\cite{fiedler-value, just-slaq}.

\section{Heat-Driven Temporal Graph Augmentation and Evolution Kernel}\label{Section-Evolution-Kernel}
\begin{figure*}[htbp]
\centering
\includegraphics[height=6cm,width=19cm]{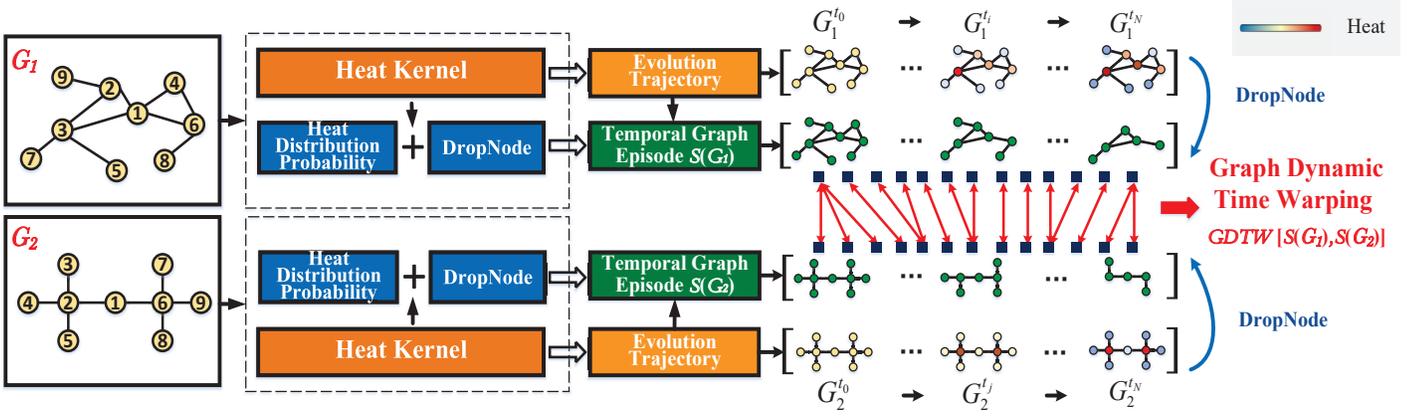}
\caption{This study demonstrates the application of Graph Dynamic Time Warping (GDTW), where two targeted graphs are represented by $G_{i}$ and $G_{j}$ respectively. Then the two temporal graph episodes are generated, and subsequently, their warping path and warping matrix are computed.}
\label{Figure-GDTW}
\end{figure*}

In this section, we proceed with the heat-driven augmentation method to convert each static graph into temporal sequence. After that, we consider the classification problem between two evolved systems and propose the graph dynamic time warping distance to measure the discrepancy between their evolutionary paths. We name this method as Evolution Kernel and show the modeling process in Figure~\ref{Figure-GDTW}.

\begin{figure*}[htbp]
\centering
\includegraphics[height=6.66cm,width=16cm]{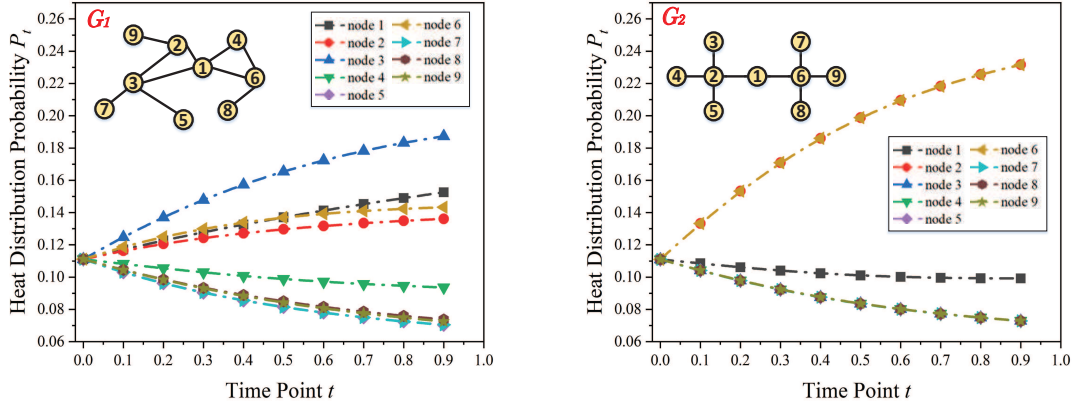}
\caption{This figure shows the heat distribution probability of each node from graph $G_{1}$ and $G_{2}$ with respect to the time point $t$. In this example, the time ranges from 0 to 0.9, and $a$, $b$ take values -2 and -2 respectively in the energy function (\ref{energy-function}). In $G_{1}$, the node $v_{3}$, $v_{1}$, $v_{6}$ and $v_{2}$ have an increasing heat distribution probabilities, and the leaf nodes $v_{5}$, $v_{7}$, $v_{8}$, $v_{9}$ have the lowest heat distribution probabilities. In $G_{2}$, the central nodes $v_{2}$ and $v_{6}$ posses the highest heat distribution probabilities, which is far greater than the bridge node $v_{1}$ and other leaf nodes.}
\label{Figure-Heat-Centrality}
\end{figure*}

\subsection{Heat-Driven Temporal Graph Augmentation}

The process of heat-driven temporal graph augmentation involves creating a sequence of augmented graphs by utilizing heat diffusion from an initial state. This technique determines the importance of each individual in a temporal graph by analyzing the heat volume of each node. Nodes with higher heat volume are considered to be more notable in an evolved system and are therefore more likely to be retained during system evolution. The augmentation of heat-driven temporal graphs comprises two essential steps: (1) establishing a correlation between the heat volume and a heat distribution probability, and (2) utilizing the heat distribution probability to produce graph augmentation.

Suppose the heat vector $U_{0}$ is initialized with the same value $u_{0}$ on each row, the solution to the equation~(\ref{heat-equation}) provides the heat at time $t$ with the form
\begin{equation}
     U_{t} = H_{t} U_{0},
\end{equation}
where the $i$-th row of $U_{t}$, denoted as $u_{t}(v_{i})$, equals the heat amount node $v_{i}$ possessing at time $t$. We calculate $U_{t}$ using the approximation techniques and analysis provided in Section~\ref{Section-Heat-Kernel}.


It may be straightforward to arrange the sequence of nodes from high to low by the volume of heat at time $t$, but here we redefine the heat distribution probability for each node by Boltzmann machine as a probability.

In the general setting, the Boltzmann machine~\cite{boltzmann-machine} for random variable $x$ is an energy-based model using an energy function,
\begin{equation}\label{Boltzmann-distribution}
  p(x) = \frac{e^{-E(x)}}{Z},
\end{equation}
where $E(x)$ denotes the energy function parameterized by weight $a$ and bias $b$ as
\begin{equation}\label{energy-function}
  E(x) = ax + b.
\end{equation}
Note that to ensure that $\sum \limits_{x}p(x) = 1$, $Z$ as the partition function is usually denoted by
\begin{equation}\label{Boltzmann-distribution}
  Z = \sum \limits_{x} e^{-E(x)}.
\end{equation}

Then we generalize the precision formulation for \textbf{heat distribution probability} with respect to node $v_{i}$ at time-point $t$, which we refer to as $P_{t}(v_{i})$, given by
\begin{equation}\label{Boltzmann-distribution}
  P_{t}(v_{i}) = \frac{e^{-E(u_{t}(v_{i}))}}{\sum \limits_{j}^{|V|} e^{-E(u_{t}(v_{j}))}},
\end{equation}
where energy function $E(x)$ is the same as the previous definition~(\ref{energy-function}).

One interesting application scenario of heat distribution probability is that we leverage it to characterize target individual's behaviors. As shown in the dolphin society example (in Figure~\ref{Figure-Dolphin-Heat-Sequence}), the intuition is that individuals with high volume energy are capable of maintaining close interactions in the complex system, while it is conceivable that ones who frequently lose energy may eventually degenerate in a way of vanishing or fading actives from the system.

To simulate these actives of each individual, we take the frequently-used graph data augmentation trick DropNode~\cite{Fastgcn} in our methodology. DropNode randomly removes a portion of nodes to create augmented data for input. Here, DropNode is utilized to simulate vanishing behaviors in each system evolution. Given a graph $G = (V, E)$ with node set $V$ and edge set $E$, and heat kernel $H_{t}$ at time-point $t$, the heat distribution probability $P_{t}(v_{i})$ of each node at time point $t$ is linked to the probability of an individual's vanishing. Precisely, a probability $\alpha_{i}$ is drawn at time $t$ from the Bernoulli distribution $\mathfrak{B}[\mathbf{norm}(P_{t}(v_{i}))]$ to determine whether a node will stay in the original system or be excluded in the next iteration, where the $\mathbf{norm}(\cdot)$ is the normalization function to uniformly normalize each node's heat distribution probability into the range of $[0, 1]$. The application of the normalized $\mathbf{norm}(P_{t}(v_{i}))$ results in the assignment of a heat distribution probability of 1 to the node with the highest heat. Consequently, the heat distribution probabilities of all other nodes are converted to a relative form, which determines their likelihood of vanishing in relation to the highest heat node. As the same time, edges connected to the removed nodes will also be deleted from the original system. The heat-driven augmented graph $G^{t} = (V^{t}, E^{t})$ at time $t$ is then obtained as described in Algorithm ~\ref{Algorithm-Node-Vanishing-Random-Graph}.

\begin{figure}[htbp]
\centering
\includegraphics[height=5.cm,width=9cm]{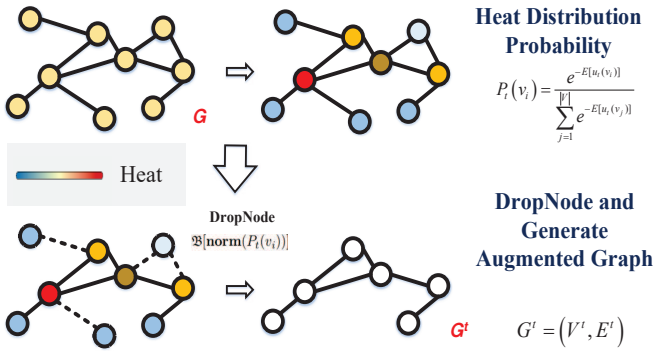}
\caption{Heat-driven temporal graph augmentation uses DropNode to generate augmented graph. We use heat distribution probability $P_{t}$ to denote the retention probability of each node at time point $t$. With $P_{t}$ to construct the Bernoulli distribution $\mathfrak{B}[\mathbf{norm}(P_{t}(v_{i}))]$, $\mathbf{DropNode}$ generates graph augmentation by removing a part of nodes and the edges connected to the dropped nodes. As a consequence, the augmented graph ${G}^{t}$ of original graph $G$ satisfies ${V}_{t}\subseteq V, {E}_{t}\subseteq E$.}
\label{Figure-DropNode}
\end{figure}

To facilitate the transition of a static graph to its dynamic evolution, we refer to the temporal graph episode as a representation for the complex system rather than relying solely on a initial graph captured at time point $t_{0}$. In particular, considering a time sequence $T = \{t_{0}, t_{1}, \ldots, t_{N}\}$, the \textbf{temporal graph episode} is given by
\begin{equation}\label{random-graph-episode}
  S(G) = \{G^{t_{0}}, G^{t_{1}}, \ldots, G^{t_{N}}\}.
\end{equation}

\begin{algorithm}
	\renewcommand{\algorithmicrequire}{\textbf{Input:}}
	\renewcommand{\algorithmicensure}{\textbf{Output:}}
	\caption{Heat-Driven Temporal Graph Augmentation}
	\label{Algorithm-Node-Vanishing-Random-Graph}
	\begin{algorithmic}[1]
		\REQUIRE graph $G = (V, E)$, time point $t$, heat distribution probability $P_{t}(v_{i})$ of each node $v_{i}$, normalization function $\mathbf{norm}(\cdot)$, Bernoulli distribution $\mathfrak{B}$.
		\ENSURE $G^{t} = (V^{t}, E^{t})$.

        \STATE initial ${V}^{t} = V$, ${E}^{t} = E$
        \FOR {$i=1$; $i<|V|$; $i++$}
            \STATE draw $\alpha_{i}$ from $\mathfrak{B}[\mathbf{norm}(P_{t}(v_{i}))]$
            \IF {$\alpha_{i}$ = 0}
                \STATE remove $v_{i}$ from set ${V}^{t}$
                \STATE remove edges connected to $v_{i}$ from ${E}^{t}$
            \ENDIF
        \ENDFOR

        \STATE \textbf{return} $G^{t} = (V^{t}, E^{t})$
	\end{algorithmic}
\end{algorithm}

\subsection{Evolution Kernel}
This part explores the distance between temporal graph episodes $S(G_{m}) = \{G_{m}^{t_{0}}, G_{m}^{t_{1}}, \ldots, G_{m}^{t_{N}}\}$ and $S(G_{n}) = \{G_{n}^{t_{0}}, G_{n}^{t_{1}}, \ldots, G_{n}^{t_{N}}\}$ and uses such a distance as the similarity measure between $G_{m}$ and $G_{n}$.

Comparing pairwise temporal graph episodes $S(G_{m})$ and $S(G_{n})$ using Euclidean distance is straightforward. The distance between two episodes is the sum of squared differences between corresponding values in time:
\begin{equation}\label{Euclidean}
  Euclidean[S(G_{m}), S(G_{n})] = [\sum_{t = t_{0}}^{t_{N}} \delta(G_{m}^{t}, G_{n}^{t})^{2}]^{\frac{1}{2}},
\end{equation}
where $\delta(G_{m}^{t}, G_{n}^{t})$ denotes the metric between $G_{m}^{t}$ and $G_{n}^{t}$. The Euclidean distance can easily be extended to multivariate temporal graph episodes, however, this distance may ignore differences in the scale of the time axis as well as shifts along the time axis. As a consequence, two temporal graph episodes that closely resemble each other from the user’s point of view might suggest a significant difference value when compared using Euclidean distance.

\begin{figure*}[htbp]
\centering
\includegraphics[height=5.5cm,width=16.5cm]{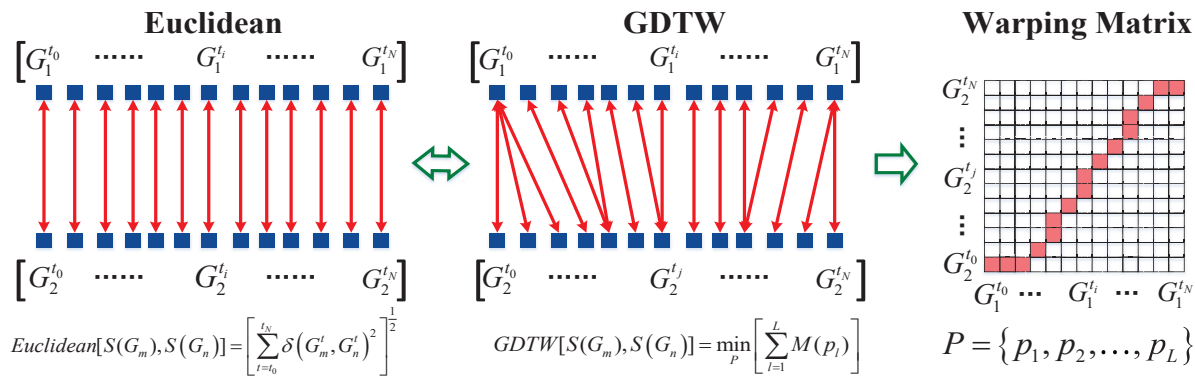}
\caption{The comparison between Euclidean distance and GDTW distance and the warping path of GDTW. Euclidean distance emphasizes the matching of corresponding points, while GDTW distance considers the alignments and shifts along the time axis. The warping path $P$ with length $L$ denotes the cell path from $p_1 = (1,1)$ to $p_L = (N, N)$ in warping matrix.}
\label{Figure-Euclidean-GDTW}
\end{figure*}

To account for this, we proposed Graph Dynamic Time Warping (GDTW), which builds on the general dynamic time warping technique outlined in~\cite{DTW,DTW-2}. By incorporating increased flexibility in scaling and shifting of the time axis, GDTW allows for more precise similarity modeling. As shown in Figure~\ref{Figure-Euclidean-GDTW}, compared to Euclidean distance, GDTW can effectively tackle the time distortion through a time-flexible alignment process that minimizes the cumulative distance between two arbitrary graph episodes. Specifically, each augmented graph within the first episode is assigned to one or more augmented graph(s) in the second episode.

For pairwise temporal graph episodes $S(G_{m})$ and $S(G_{n})$, the $N \times N$ warping matrix $\mathbf{M}$ is constructed with $(i, j)$ element $\mathbf{M}(i, j)$ denoting the metric $\delta(G_{m}^{t_{i}}, G_{n}^{t_{j}})$, which corresponds to the alignment between $G_{m}^{t_{i}}$ and $G_{n}^{t_{j}}$.

The warping path is a continuous sequence of matrix cells that defines a mapping between temporal graph episodes $S(G_{m})$ and $S(G_{n})$. To find the optimal alignment between $S(G_{m})$ and $S(G_{n})$, we can find a warping path through the matrix that minimizes the total cumulative value of all elements it has passed.

Let $P = \{ p_{1}, p_{2}, \ldots, p_{L}\}$ denotes the \textbf{warping path} with length $L$, where $p_{l} = (i, j)$ denotes the alignment between $G_{m}^{t_{i}}$ and $G_{n}^{t_{j}}$.
Here the alignment satisfies the following three constraints:
\begin{itemize}
\item[(1)] Boundary: $p_{1} = (1, 1)$ and $p_{L} = (N, N)$.
\item[(2)] Monotonicity: If $p_{l} = (i, j)$ and $p_{l+1} = (i', j')$, then $i' \geq i$ and $j' \geq j$.
\item[(3)] Continuity: If $p_{l} = (i, j)$ and $p_{l+1} = (i', j')$, then $i' \leq i+1$ and $j' \leq j+1$.
\end{itemize}

Then the GDTW distance is defined as the minimal cost of possible matches between $S(G_{m})$ and $S(G_{n})$:
\begin{equation}\label{GDTW}
  GDTW[S(G_{m}), S(G_{n})] = \min\limits_{P}[\sum_{l = 1}^{L} \mathbf{M}(p_{l})],
\end{equation}
where $p_{l} = (i, j)$ and $\mathbf{M}(p_{l})= \delta(G_{m}^{t_{i}}, G_{n}^{t_{j}})$.

The GDTW distance can be recursively calculated using dynamic programming~\cite{dynamic-programming} based on the cumulative distance $\gamma(i, j)$, which denotes the sum of the metric $\mathbf{M}(i, j)$ in current $(i, j)$ position and the minimum among the previous cumulative distance of the adjacent positions:
\begin{equation}\label{GDTW-DP}
  \gamma(i, j) = \mathbf{M}(i, j) + \min\{ \gamma(i, j-1), \gamma(i-1, j), \gamma(i-1, j-1) \},
\end{equation}
where $\gamma(0, 0) = 0$, $\gamma(i, 0) = \infty$, $\gamma(0, j) = \infty$, for $i, j \geq 1$.

In the last, we explain the metric definition $\delta(G_{m}^{t_{i}}, G_{n}^{t_{j}})$ in formula~(\ref{GDTW}). As structural data, graphs cannot be not directly calculated for pairwise distance in $\delta(G_{m}^{t_{i}}, G_{n}^{t_{j}})$. Therefore, our approach transforms each graph into its continuous vectorial embedding form to represent the structural data. To achieve this, we utilize the graph2vec~\cite{graph2vec} technique for graph embedding, and take Euclidean distance to calculate pairwise distance $\delta(G_{m}^{t_{i}}, G_{n}^{t_{j}})$ in this study.

In our paper, the \textbf{Evolution kernel} method refers to the utilization of the calculated GDTW distances as the basis for the kernel function incorporated in classifiers such as the Support Vector Machine.

\begin{algorithm}
	\renewcommand{\algorithmicrequire}{\textbf{Input:}}
	\renewcommand{\algorithmicensure}{\textbf{Output:}}
	\caption{Graph Dynamic Time Warping}
	\label{Algorithm-Graph-Dynamic-Time-Warping}
	\begin{algorithmic}[1]
		\REQUIRE $S(G_{m}) = \{G_{m}^{t_{0}}, G_{m}^{t_{1}}, \ldots, G_{m}^{t_{N}}\}$ and $S(G_{n}) = \{G_{n}^{t_{0}}, G_{n}^{t_{1}}, \ldots, G_{n}^{t_{N}}\}$.
		\ENSURE GDTW distance $GDTW[S(G_{m}), S(G_{n})]$.

        \STATE initiate the $N \times N$ warping matrix $\mathbf{M}$ with $\mathbf{M}(i, j) = \delta(G_{m}^{t_{i}}, G_{n}^{t_{j}})$,
        \STATE initiate the cumulative distance $\gamma(i, j)$ with $\gamma(0, 0) = 0$, $\gamma(i, 0) = \infty$, $\gamma(0, j) = \infty$, for $i, j \geq 1$.

        \FOR {$i=1$; $i<N$; $i++$}
            \FOR {$j=1$; $j<N$; $j++$}
                \STATE $\gamma(i, j) = \mathbf{M}(i, j) + \min\{ \gamma(i, j-1), \gamma(i-1, j), \gamma(i-1, j-1) \}$,
            \ENDFOR
        \ENDFOR

        \STATE \textbf{return} $GDTW[S(G_{m}), S(G_{n})] = \gamma(N, N)$.
	\end{algorithmic}
\end{algorithm}

\section{Experiments}\label{Section-Experiments}
With the proposed model above, we evaluate the effectiveness of our model on the supervised graph classification task in this section. In the supervised graph classification setting, each graph is assigned a label from a finite set $\mathcal{L}$. During the training, a part of graphs with labels are treated as a training set, and the task is to predict the labels for the remaining graphs.

We begin by introducing the datasets for experiments and the baselines for comparison, then we present the outcomes of our evolution kernel model and analyze the evolutionary time $t$.

\subsection{Dataset}
We evaluate our proposed method using two categories of real-world graph datasets, i.e., molecule datasets and social network datasets, whose statistics are summarized in Table~\ref{dataset-molecule}, and Table~\ref{dataset-social-network}. Their details are as follows.
\subsubsection{Molecule Datasets}
\begin{itemize}
  \item \textbf{MUTAG}~\cite{dataset-mutag} is a dataset of aromatic and heteroaromatic nitro compounds labeled according to whether they have a mutagenic effect on bacteria or not.
  \item \textbf{DD}~\cite{dataset-dd} is a dataset of protein structures where nodes represent amino acids and edges indicate spatial closeness, which are classiﬁed into enzymes or non-enzymes.
  \item \textbf{PTC-MR}~\cite{dataset-ptc-mr} consists of graph representations of chemical molecules labeled according to carcinogenicity on rodents.
  \item \textbf{PROTEINS}~\cite{dataset-protein} is a set of protein graphs with nodes representing secondary structure elements and edges indicating neighborhood in the amino-acid sequence or in 3-dimension space.
  \item \textbf{ENZYMES}~\cite{dataset-protein} consists of protein tertiary structures obtained from the BRENDA enzyme database.
  \item \textbf{NCI-1}, \textbf{NCI-109}~\cite{dataset-nci} are datasets of chemical compounds divided by the anti-cancer property (active or negative). The National Cancer Institute (NCI) had made these datasets publicly available.
\end{itemize}

\begin{table}[htbp]
  \centering
  \caption{The molecule datasets' statistics include graph classes, graph number, average node number, and average edge number.}
    \begin{tabular}{lcccc}
    \toprule
    \toprule
    \multicolumn{1}{c}{\multirow{2}[0]{*}{\textbf{Dataset}}} & \textbf{Class} & \textbf{Graph} & \textbf{Average Node} & \textbf{Average Edge} \\
          \cline{2-5} & $\mathcal{C}$     & $N$     & $\bar{|V|}$     & $\bar{|E|}$ \\
    \midrule
    MUTAG~\cite{dataset-mutag}          & 2         & 188       & 17.93             & 19.79\\
    DD~\cite{dataset-dd}                & 2         & 1178      & 284.32            & 715.66\\
    PTC-MR~\cite{dataset-ptc-mr}        & 2         & 344       & 14.29             & 14.69\\
    PROTEINS~\cite{dataset-protein}     & 2         & 1113      & 39.06             & 72.82\\
    ENZYMES~\cite{dataset-protein}      & 6         & 600       & 32.63             & 64.14\\
    NCI-1~\cite{dataset-nci}            & 2         & 4110      & 29.87             & 32.30\\
    NCI-109~\cite{dataset-nci}          & 2         & 4127      & 29.68             & 32.13\\
    \bottomrule
    \end{tabular}%
  \label{dataset-molecule}%
\end{table}%

\subsubsection{Social Network Datasets}
\begin{itemize}
  \item \textbf{IMDB-BINARY}~\cite{dataset-mutag} is a movie collaboration dataset sourced from IMDB. The dataset consists of ego-network graphs, with each graph representing actors and actresses as nodes and edges indicating their co-appearance in a movie. The dataset is further classified into two distinct genres: Action and Romance.
  \item \textbf{COLLAB}~\cite{dataset-dd} represents a scientific collaboration network. Each graph in the dataset depicts an ego network of a researcher, wherein both the researcher and their collaborators are denoted as nodes, and the edges depict the collaboration between two researchers. The researcher's ego network is assigned one of three labels - High Energy Physics, Condensed Matter Physics, and Astro Physics - based on the field to which they belong.
  \item \textbf{Reddit-Binary}~\cite{dataset-ptc-mr} consists of graphs corresponding to online discussions on Reddit. In each graph, nodes represent users; there is an edge between them if at least one of them responds to the other’s comment. A graph is labeled according to whether it belongs to a question/answer-based community or a discussion-based community.
\end{itemize}

\begin{table}
\centering
\caption{The social network datasets' statistics include graph classes, graph number, average node number, and average edge number.}
{
\begin{tabular}{p{47 pt}cccc}
\toprule
\toprule
\multicolumn{1}{c}{\multirow{2}[0]{*}{\textbf{Dataset}}} & \textbf{Class} & \textbf{Graph} & \textbf{Average Node} & \textbf{Average Edge} \\
          \cline{2-5} & $\mathcal{C}$     & $N$     & $\bar{|V|}$     & $\bar{|E|}$ \\
\midrule
IMDB-Binary~\cite{dataset-mutag}         & 2                 & 1000              & 19.77                 & 193.06\\
COLLAB~\cite{dataset-dd}                 & 2                 & 5000              & 74.49                 & 4914.99\\
Reddit-Binary~\cite{dataset-ptc-mr}      & 2                 & 2000              & 429.61                & 193.06\\
\bottomrule
\end{tabular}}
\label{dataset-social-network}
\end{table}

\subsection{Baselines}
To validate the performance of our approach, we compare it with graph embedding methods, graph neural networks, and graph augmentation methods on datasets in Table~\ref{Table-results-molecule} and Table~\ref{Table-results-social}. We categorize these competitors into four main groups and the details of the learning methods that are compared.

\begin{itemize}
    \item \textbf{Kernel methods}. \textbf{Weisfeiler-Lehman kernel (WL)}~\cite{WL-kernel} maps graph data into a Weisfeiler-Lehman sequence, whose node attributes represent graph topology and label information. \textbf{Graphlet kernel (GK)}~\cite{GK-graphlet-kernel} measures graph similarity by counting common $k$-node graphlets. \textbf{Deep Weisfeiler-Lehman (Deep WL), Deep Graphlet (Deep GK)}~\cite{Deep-graph-kernels}: Deep Weisfeiler-Lehman kernel and Deep Graphlet kernel leverage language modeling and deep learning to learn latent representations of sub-structures for graphs. \textbf{Persistence Fisher} kernel~\cite{Persistence-Fisher} relies on Fisher information geometry to explore persistence diagrams on structural graph pattern recognition. \textbf{WKPI}~\cite{WKPI}: WKPI designs a weighted kernel for persistence images, together with a metric-learning method to learn the best classify function for labeled data.
    \item \textbf{Embedding methods}. \textbf{Graph2vec}~\cite{graph2vec} treats rooted subgraphs as words and graphs as sentences or documents, and then it uses Skip-gram in Natural Language Processing to get explicit graph embeddings. \textbf{AWE}~\cite{AWE} uses anonymous random walks to embed entire graphs in an unsupervised manner, but it takes a different embedding strategy compared with our methodology. AWE leverages the neighborhoods of anonymous walks while our work focuses on the co-occurring anonymous walks on a global scale.
    \item \textbf{Graph Neural Networks methods}. Analogous to convolutional neural networks, \textbf{PATCHY-SAN}~\cite{PSCN} proposes a framework to perform convolutional operations for arbitrary graph data. \textbf{GCN} (Graph Convolutional Networks)~\cite{GCN} proposes convolutional architecture via a local first-order approximation containing both local graph structure and features of nodes.
\end{itemize}

\subsection{Experimental Results}
\begin{figure}[htbp]
\centering
\includegraphics[height=7cm,width=9cm]{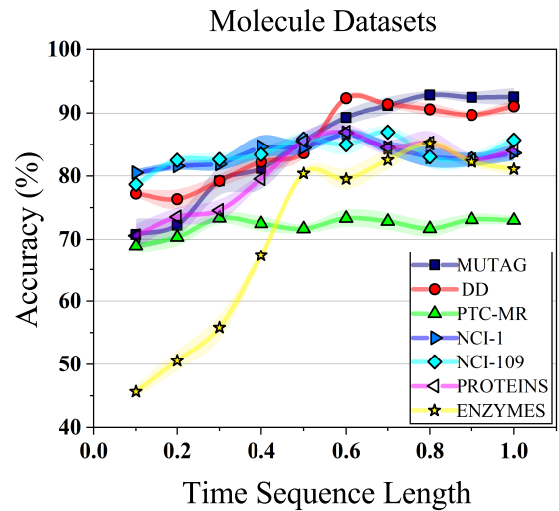}
\caption{The time grows from 0.1 to 1 with time interval 0.1. We use Suppert Vector Machine as the classifier and show the supervised classification on molecule datasets.
The shadow indicts the standard deviation derived by 10-fold cross validation.}
\label{Figure-Result-molecule}
\end{figure}

The classification accuracy results for labeled molecule datasets are presented in Table~\ref{Table-results-molecule} and Figure~\ref{Figure-Result-molecule}. The accuracy was obtained through the 10-fold cross-validation method, and the corresponding standard deviations were reported.

Table~\ref{Table-results-molecule} provides the molecule datasets' results. In molecule datasets, we regard the decomposition of molecular structures as molecules' evolution. In Table~\ref{Table-results-molecule}, our method consistently ranks at the top for seven datasets. Remarkably, our methodology demonstrated significant improvement in predicted accuracy for DD, NCI-1, NCI-109, and ENZYMES, in comparison to kernel methods, embedding methods, and graph neural network methods, achieving 92.33\%, 86.85\%, 86.95\%, and 85.20\%, respectively. In Figure~\ref{Figure-Result-molecule}, we set a time interval as 0.1 and take the time length for heat kernel evolution to 1. The overall results are positively associated with the time series length for most datasets, including MUTAG, DD, NCI-1, NCI-109, PROTEINS, and ENZYMES. Specifically, ENZYMES demonstrats greater sensitivity to the time length, while PTC-MR shows less fluctuation.

\begin{figure}[htbp]
\centering
\includegraphics[height=7cm,width=9cm]{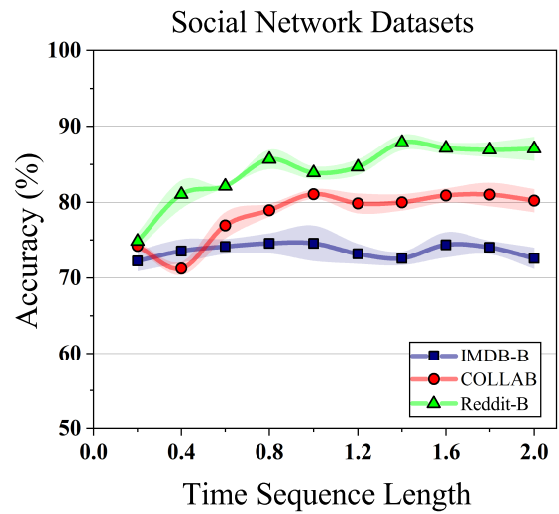}
\caption{The time grows from 0.2 to 2 with time interval 0.2. We use Suppert Vector Machine as the classifier and show the supervised classification on social network datasets. The shadow indicts the standard deviation derived by 10-fold cross validation.}
\label{Figure-Result-social}
\end{figure}

The outcomes of our experiment on classification accuracy for labeled social network datasets are presented in Table~\ref{Table-results-social} and Figure~\ref{Figure-Result-social}.

In social networks, we take the interaction between individuals and the retention or disappearance of individuals as the evolution of the system. As social networks are typically comprised of numerous elements and interactions, we opt to conduct a longer evolution trajectory with a time length 2 and a time interval of 0.2. In Table~\ref{Table-results-social}, our approach still promotes the classification in three datasets. Notably, in Figure~\ref{Figure-Result-social}, our method displays favorable classification effectiveness as time progressed. Conversely, accuracy seems relatively insensitive to evolution for the IMDB-B social network, as the performance fluctuation remains relatively stable along the time axis.

According to the analysis, it is evident that the classification results for the datasets can be largely attributed to the evolution time of the heat kernel. A long time is advantageous in identifying disparities among the systems displayed on a graph while simultaneously amplifying the gap between their evolutionary trajectories.

\begin{table*}
\centering
\caption{Classiﬁcation accuracy (standard deviation) of our method and state-of-the-art baselines on molecule datasets. “-” means the classiﬁcation accuracy (standard deviation) is not available in the original papers. We mark the best performance in a column in bold font.
}
{
\begin{tabular}{lllllllllllll}
\toprule
\textbf{Algorithm}                          & \textbf{MUTAG} & \textbf{DD} & \textbf{PTC-MR} & \textbf{NCI-1} & \textbf{NCI-109} & \textbf{PROTEINS} & \textbf{ENZYMES} \\
\midrule
WL~\cite{weisfeiler-lehman-graph-kernels}   & 80.63 (3.07)& 77.95 (0.70)& 56.97 (2.01)& 80.13 (0.50)& 80.22 (0.34)& 72.92 (0.56)& 53.15 (1.14)\\
Deep WL~\cite{deep-graph-kernel}            & 82.95 (1.96)& -           & 59.17 (1.56)& \textcolor{blue}{80.31 (0.46)}& 80.32 (0.33)& 73.30 (0.82)& \textcolor{blue}{53.43 (0.91)}\\
GK~\cite{graphlet-kernel}                   & 81.66 (2.11)& 78.45 (0.26)& 57.26 (1.41)& 62.28 (0.29)& 62.60 (0.19)& 71.67 (0.55)& 26.61 (0.99)\\
Deep GK~\cite{deep-graph-kernel}            & 82.66 (1.45)& -           & 57.32 (1.13)& 62.48 (0.25)& 62.69 (0.23)& 71.68 (0.50)& 27.08 (0.79)\\
Persistence Fisher~\cite{Persistence-Fisher}& 85.60 (1.70)& 79.40 (0.80)& 62.42 (1.80)&-            &-            & 75.20 (2.10)& -\\
WKPI~\cite{WKPI}                            & 85.80 (2.50)& \textcolor{blue}{82.00 (0.50)}& 62.70 (2.70)&-            &-            & \textcolor{blue}{78.50 (0.40)}& -\\
graph2vec~\cite{node2vec}                   & 83.15 (9.25)& 58.64 (0.01)& 60.17 (6.86)& 73.22 (1.81)& 74.26 (1.47)& 73.30 (2.05)& 44.33 (0.09)\\
AWE~\cite{AWE}                              & 87.87 (9.76)& 71.51 (4.02)& 59.14 (1.83)& 62.72 (1.67)& 63.21 (1.42)& 70.01 (2.52)& 35.77 (5.93)\\
PSCN~\cite{PSCN}                            & \textcolor{blue}{92.63 (4.21)}& 77.12 (2.41)& 60.00 (4.82)& 78.59 (1.89)& -           & 75.89 (2.76)& -\\
GCN~\cite{GCN}                              & 91.64 (7.20)& 66.83 (4.30)& \textcolor{blue}{71.35 (0.64)}& 76.27 (4.10)& \textcolor{blue}{80.47 (0.19)}& 67.21 (3.00)& 24.26 (4.70)\\
\midrule
\textbf{Evolution Kernel}                   & \textbf{92.90 (1.06)}& \textbf{92.33(0.52)}& \textbf{73.52 (0.13)}& \textbf{86.85 (0.09)}& \textbf{86.95 (0.23)}& \textbf{83.69 (0.11)}& \textbf{85.20 (0.74)}\\
\bottomrule
\end{tabular}}
\label{Table-results-molecule}
\end{table*}

\begin{table}
\centering
\caption{Classiﬁcation accuracy (standard deviation) of our method and state-of-the-art baselines on social network datasets. “-” means the classiﬁcation accuracy (standard deviation) is not available in the original papers. We mark the best result in a column in bold font.}
{
\begin{tabular}{lllllllllllll}
\toprule
\textbf{Algorithm}                          & \textbf{IMDB-Binary}               & \textbf{COLLAB}                   & \textbf{Reddit-Binary}         \\
\midrule
WL~\cite{weisfeiler-lehman-graph-kernels}   & 73.40 (4.63)                  & \textcolor{blue}{79.02 (1.77)}    & 81.10 (1.90)              \\
GK~\cite{graphlet-kernel}                   & 65.87 (0.98)                  & 72.84 (0.28)                      & 65.87 (0.98)              \\
Deep GK~\cite{deep-graph-kernel}            & 66.96 (0.56)                  & 73.09 (0.25)                      & 78.04 (0.39)              \\
graph2vec~\cite{node2vec}                   & 71.00 (2.29)                  & -                                 & -                         \\
AWE~\cite{AWE}                              & \textcolor{blue}{74.45 (5.83)}& 73.93 (1.94)                      & \textcolor{blue}{87.89 (2.53)}  \\
PSCN~\cite{PSCN}                            & 71.00 (2.29)                  & 72.60 (2.15)                      & 86.30 (1.58)              \\
\midrule
\textbf{Evolution Kernel}                   & \textbf{74.56(1.33)}          & \textbf{81.08(0.24)}              & \textbf{88.97 (0.87)}     \\
\bottomrule
\end{tabular}}
\label{Table-results-social}
\end{table}

\section{Conclusion}
Autonomous individuals establish a structural system by means of different connections. With individuals interacting over the time axis, the system sprouts evolution and then displays evolution from statistical or sociological perspectives. The evolutional trajectory highlights the natural principles of evolutionary dynamics, thereby describing the heterogeneity of a system.

This research paper presents a unique idea for the classification of static graphs. Rather than relying solely on their static forms, we propose an approach by comparing two graphs based on their substantial evolutionary dynamics. To achieve this, we employ the physics-based dynamic evolution model, the heat kernel, to simulate the information propagation within a system. Our approach involves creating a heat distribution probability to accurately capture the evolutionary dynamics of each individual in a static graph. This is done by assigning heat distribution probability weight to each point in the graph at a moment and using DropNode to simulate elements' retention or disappearance from the system. To assess the distance between pairwise evolutionary systems, we developed the dynamic time wrapping distance GDTW with respect to the time series through optimal matching. Our experiments demonstrate that our method using SVM classifier with an evolution kernel significantly enhances classification performance across a range of molecule and social network datasets. In particular, the classification accuracies can be improved by considering graphs' evolutionary behaviors, rather than relying solely on their static forms. Additionally, we found that the length of the time sequence is essential for achieving optimal classification outcomes.

For future work, several interesting problems need to be studied further. We tend to research additional mechanisms to reveal various evolutionary dynamics in diverse systems. These mechanisms include the evolution through pinning control, as well as the evolution through self-organizing, etc. Additionally, we intend to investigate notable graph indexes that can assist in revealing the evolutionary attributes in geometry, such as the eigenspectrum, which would allow us to better understand the structural evolution of graph models.

\section*{Acknowledgements}
This work was supported by the National Natural Science Foundation of China (Grant Nos. 62276013, 62141605, 62050132), the Beijing Natural Science Foundation (Grant No. 1192012), and the Fundamental Research Funds for the Central Universities.

\bibliographystyle{unsrt}
\bibliography{graph_dynamics.bib}

\begin{thebibliography}{10}

\bibitem{network-social-animal}
Josefine~Bohr Brask, Samuel Ellis, and Darren~P Croft.
\newblock Animal social networks: an introduction for complex systems
  scientists.
\newblock {\em Journal of Complex Networks}, 9(2):cnab001, 2021.
\newblock
  doi:{\color{blue}\href{https://doi.org/10.1093/comnet/cnab001}{10.1093/comnet/cnab001}}.

\bibitem{network-Social-Dolphin}
Rebecca~A Hamilton, Teresa Borcuch, Simon~J Allen, William~R Cioffi, Vanni
  Bucci, Michael Kr{\"u}tzen, and Richard~C Connor.
\newblock Male aggression varies with consortship rate and habitat in a dolphin
  social network.
\newblock {\em Behavioral Ecology and Sociobiology}, 73:1--7, 2019.
\newblock
  doi:{\color{blue}\href{https://doi.org/10.1007/s00265-019-2753-1}{10.1007/s00265-019-2753-1}}.

\bibitem{network-social-opinion}
Mayte P{\'e}rez-Llanos, Juan~Pablo Pinasco, and Nicolas Saintier.
\newblock Opinion attractiveness and its effect in opinion formation models.
\newblock {\em Physica A: Statistical Mechanics and its Applications},
  559:125017, 2020.
\newblock
  doi:{\color{blue}\href{https://doi.org/10.1016/j.physa.2020.125017}{10.1016/j.physa.2020.125017}}.

\bibitem{network-WWW}
Robert Cooley, Bamshad Mobasher, and Jaideep Srivastava.
\newblock Data preparation for mining world wide web browsing patterns.
\newblock {\em Knowledge and information systems}, 1(1):5--32, 1999.
\newblock
  doi:{\color{blue}\href{https://doi.org/10.1007/BF03325089}{10.1007/BF03325089}}.

\bibitem{pagerank}
Mehdi Joodaki, Mohammad~Bagher Dowlatshahi, and Nazanin~Zahra Joodaki.
\newblock An ensemble feature selection algorithm based on pagerank centrality
  and fuzzy logic.
\newblock {\em Knowledge-Based Systems}, 233:107538, 2021.
\newblock
  doi:{\color{blue}\href{https://doi.org/10.1016/j.knosys.2021.107538}{10.1016/j.knosys.2021.107538}}.

\bibitem{WL-kernel}
Dai~Hai Nguyen, Canh~Hao Nguyen, and Hiroshi Mamitsuka.
\newblock Learning subtree pattern importance for weisfeiler-lehman based graph
  kernels.
\newblock {\em Machine Learning}, 110(7):1585--1607, 2021.
\newblock
  doi:{\color{blue}\href{https://doi.org/10.1007/s10994-021-05991-y}{10.1007/s10994-021-05991-y}}.

\bibitem{GK-graphlet-kernel}
Furqan Aziz, Afan Ullah, and Faiza Shah.
\newblock Feature selection and learning for graphlet kernel.
\newblock {\em Pattern Recognition Letters}, 136:63--70, 2020.
\newblock
  doi:{\color{blue}\href{https://doi.org/10.1016/j.patrec.2020.05.023}{10.1016/j.patrec.2020.05.023}}.

\bibitem{Deep-graph-kernels}
Pinar Yanardag and SVN Vishwanathan.
\newblock Deep graph kernels.
\newblock In {\em Proceedings of the 21th ACM SIGKDD international conference
  on knowledge discovery and data mining}, pages 1365--1374, 2015.
\newblock
  doi:{\color{blue}\href{https://doi.org/10.1145/2783258.2783417}{10.1145/2783258.2783417}}.

\bibitem{Persistence-Fisher}
Tam Le and Makoto Yamada.
\newblock Persistence fisher kernel: A riemannian manifold kernel for
  persistence diagrams.
\newblock {\em Advances in Neural Information Processing Systems}, 31, 2018.
\newblock
  doi:{\color{blue}\href{https://doi.org/10.48550/arXiv.1802.03569}{10.48550/arXiv.1802.03569}}.

\bibitem{WKPI}
Qi~Zhao and Yusu Wang.
\newblock Learning metrics for persistence-based summaries and applications for
  graph classification.
\newblock {\em Advances in Neural Information Processing Systems}, 32, 2019.
\newblock
  doi:{\color{blue}\href{https://doi.org/10.48550/arXiv.1904.12189}{10.48550/arXiv.1904.12189}}.

\bibitem{heat-diffusion}
Haotian Zhao, Bin Wang, Haotian Liu, Hongbin Sun, Zhaoguang Pan, and Qinglai
  Guo.
\newblock Exploiting the flexibility inside park-level commercial buildings
  considering heat transfer time delay: A memory-augmented deep reinforcement
  learning approach.
\newblock {\em IEEE Transactions on Sustainable Energy}, 13(1):207--219, 2022.
\newblock
  doi:{\color{blue}\href{http://doi.org/10.1109/TSTE.2021.3107439}{10.1109/TSTE.2021.3107439}}.

\bibitem{heat-kernel-community-detection}
Kyle Kloster and David~F Gleich.
\newblock Heat kernel based community detection.
\newblock In {\em Proceedings of the 20th ACM SIGKDD international conference
  on Knowledge discovery and data mining}, pages 1386--1395, 2014.
\newblock
  doi:{\color{blue}\href{https://doi.org/10.1145/2623330.2623706}{10.1145/2623330.2623706}}.

\bibitem{Heat-Kernel-diffusionrank}
Haixuan Yang, Irwin King, and Michael~R Lyu.
\newblock Diffusionrank: a possible penicillin for web spamming.
\newblock In {\em Proceedings of the 30th annual international ACM SIGIR
  conference on Research and development in information retrieval}, pages
  431--438, 2007.
\newblock
  doi:{\color{blue}\href{https://doi.org/10.1145/1277741.1277815}{10.1145/1277741.1277815}}.

\bibitem{spectral-graph-theory}
Fan~RK Chung and Fan~Chung Graham.
\newblock {\em Spectral graph theory}.
\newblock Number~92. American Mathematical Soc., 1997.

\bibitem{GCN}
Thomas~N Kipf and Max Welling.
\newblock Semi-supervised classification with graph convolutional networks.
\newblock In {\em Proceedings International Conference on Learning
  Representations}, 2017.
\newblock
  doi:{\color{blue}\href{https://doi.org/10.48550/arXiv.1609.02907}{10.48550/arXiv.1609.02907}}.

\bibitem{heat-kernel}
Shih-Gu Huang, Ilwoo Lyu, Anqi Qiu, and Moo~K Chung.
\newblock Fast polynomial approximation of heat kernel convolution on manifolds
  and its application to brain sulcal and gyral graph pattern analysis.
\newblock {\em IEEE transactions on medical imaging}, 39(6):2201--2212, 2020.
\newblock
  doi:{\color{blue}\href{https://doi.org/10.1109/TMI.2020.2967451}{10.1109/TMI.2020.2967451}}.

\bibitem{diffusion-kernels}
Risi~Imre Kondor and John Lafferty.
\newblock Diffusion kernels on graphs and other discrete structures.
\newblock In {\em Proceedings of the 19th international conference on machine
  learning}, volume 2002, pages 315--322, 2002.

\bibitem{sun2009concise-heat-diffusion}
Jian Sun, Maks Ovsjanikov, and Leonidas Guibas.
\newblock A concise and provably informative multi-scale signature based on
  heat diffusion.
\newblock In {\em Computer graphics forum}, volume~28, pages 1383--1392. Wiley
  Online Library, 2009.
\newblock
  doi:{\color{blue}\href{https://doi.org/10.1111/j.1467-8659.2009.01515.x}{10.1111/j.1467-8659.2009.01515.x}}.

\bibitem{matrix_sensitivity}
Charles Van~Loan.
\newblock The sensitivity of the matrix exponential.
\newblock {\em SIAM Journal on Numerical Analysis}, 14(6):971--981, 1977.
\newblock
  doi:{\color{blue}\href{https://doi.org/10.1137/0714065}{10.1137/0714065}}.

\bibitem{fiedler-value}
Miroslav Fiedler.
\newblock A property of eigenvectors of nonnegative symmetric matrices and its
  application to graph theory.
\newblock {\em Czechoslovak mathematical journal}, 25(4):619--633, 1975.

\bibitem{just-slaq}
Anton Tsitsulin, Marina Munkhoeva, and Bryan Perozzi.
\newblock Just slaq when you approximate: accurate spectral distances for
  web-scale graphs.
\newblock In {\em Proceedings of the Web Conference 2020}, pages 2697--2703,
  2020.
\newblock
  doi:{\color{blue}\href{https://doi.org/10.1145/3366423.3380026}{10.1145/3366423.3380026}}.

\bibitem{boltzmann-machine}
Roger~G. Melko, Giuseppe Carleo, Juan Carrasquilla, and J.~Ignacio Cirac.
\newblock Restricted boltzmann machines in quantum physics.
\newblock {\em Nature Physics}, 15:887--892, 2019.
\newblock
  doi:{\color{blue}\href{http://doi.org/10.1038/s41567-019-0545-1}{10.1038/s41567-019-0545-1}}.

\bibitem{Fastgcn}
Jie Chen, Tengfei Ma, and Cao Xiao.
\newblock Fastgcn: fast learning with graph convolutional networks via
  importance sampling.
\newblock In {\em Proceedings International Conference on Learning
  Representations}, 2018.
\newblock
  doi:{\color{blue}\href{https://doi.org/10.48550/arXiv.1801.10247}{10.48550/arXiv.1801.10247}}.

\bibitem{DTW}
Huanhuan Li, Jingxian Liu, Zaili Yang, Ryan~Wen Liu, Kefeng Wu, and Yuan Wan.
\newblock Adaptively constrained dynamic time warping for time series
  classification and clustering.
\newblock {\em Information Sciences}, 534:97--116, 2020.
\newblock
  doi:{\color{blue}\href{https://doi.org/10.1016/j.ins.2020.04.009}{10.1016/j.ins.2020.04.009}}.

\bibitem{DTW-2}
Time-series averaging and local stability-weighted dynamic time warping for
  online signature verification.
\newblock {\em Pattern Recognition}, 112:107699, 2021.
\newblock
  doi:{\color{blue}\href{http://doi.org/10.1016/j.patcog.2020.107699}{10.1016/j.patcog.2020.107699}}.

\bibitem{dynamic-programming}
Derong Liu, Shan Xue, Bo~Zhao, Biao Luo, and Qinglai Wei.
\newblock Adaptive dynamic programming for control: A survey and recent
  advances.
\newblock {\em IEEE Transactions on Systems, Man, and Cybernetics: Systems},
  51(1):142--160, 2020.
\newblock
  doi:{\color{blue}\href{https://doi.org/10.1109/TSMC.2020.3042876}{10.1109/TSMC.2020.3042876}}.

\bibitem{graph2vec}
Annamalai Narayanan, Mahinthan Chandramohan, Rajasekar Venkatesan, Lihui Chen,
  Yang Liu, and Shantanu Jaiswal.
\newblock graph2vec: Learning distributed representations of graphs.
\newblock {\em arXiv preprint arXiv:1707.05005}, 2017.
\newblock
  doi:{\color{blue}\href{https://doi.org/10.48550/arXiv.1707.05005}{10.48550/arXiv.1707.05005}}.

\bibitem{dataset-mutag}
Asim~Kumar Debnath, Rosa L. Lopez~De Compadre, Gargi Debnath, Alan~J.
  Shusterman, and Corwin Hansch.
\newblock Structure-activity relationship of mutagenic aromatic and
  heteroaromatic nitro compounds. correlation with molecular orbital energies
  and hydrophobicity.
\newblock {\em Journal of Medicinal Chemistry}, 34(2):786--797, 1991.
\newblock
  doi:{\color{blue}\href{https://doi.org/10.1021/jm00106a046}{10.1021/jm00106a046}}.

\bibitem{dataset-dd}
Paul~D. Dobson and Andrew~J. Doig.
\newblock Distinguishing enzyme structures from non-enzymes without alignments.
\newblock {\em Journal of Molecular Biology}, 330(4):771--783, 2003.
\newblock
  doi:{\color{blue}\href{https://doi.org/10.1016/S0022-2836(03)00628-4}{10.1016/S0022-2836(03)00628-4}}.

\bibitem{dataset-ptc-mr}
C.~Helma, R.~D. King, S.~Kramer, and A.~Srinivasan.
\newblock The predictive toxicology challenge 2000-2001.
\newblock {\em Bioinformatics}, 17(1):107--108, 2001.
\newblock
  doi:{\color{blue}\href{https://doi.org/10.1093/bioinformatics/17.1.107}{10.1093/bioinformatics/17.1.107}}.

\bibitem{dataset-protein}
Karsten~M. Borgwardt, Ong~Cheng Soon, Schönauer Stefan, S.~V.~N. Vishwanathan,
  Alex~J. Smola, and Kriegel Hans-Peter.
\newblock Protein function prediction via graph kernels.
\newblock {\em Bioinformatics}, 21(suppl\_1):i47--i56, 2005.
\newblock
  doi:{\color{blue}\href{https://doi.org/10.1093/bioinformatics/bti1007}{10.1093/bioinformatics/bti1007}}.

\bibitem{dataset-nci}
John~Boaz Lee, Ryan Rossi, and Xiangnan Kong.
\newblock Graph classification using structural attention.
\newblock In {\em Proceedings of the 24th ACM SIGKDD International Conference
  on Knowledge Discovery \&amp; Data Mining}, pages 1666--1674, 2018.
\newblock
  doi:{\color{blue}\href{https://doi.org/10.1145/3219819.3219980}{10.1145/3219819.3219980}}.

\bibitem{AWE}
Sergey Ivanov and Evgeny Burnaev.
\newblock Anonymous walk embeddings.
\newblock In {\em International conference on machine learning}, pages
  2186--2195. PMLR, 2018.
\newblock
  doi:{\color{blue}\href{https://doi.org/10.48550/arXiv.1805.11921}{10.48550/arXiv.1805.11921}}.

\bibitem{PSCN}
Mathias Niepert, Mohamed Ahmed, and Konstantin Kutzkov.
\newblock Learning convolutional neural networks for graphs.
\newblock In {\em International conference on machine learning}, pages
  2014--2023. PMLR, 2016.
\newblock
  doi:{\color{blue}\href{https://doi.org/10.48550/arXiv.1605.05273}{10.48550/arXiv.1605.05273}}.

\bibitem{weisfeiler-lehman-graph-kernels}
Nino Shervashidze, Pascal Schweitzer, Erik Jan, Van Leeuwen, and Karsten~M.
  Borgwardt.
\newblock Weisfeiler-lehman graph kernels.
\newblock {\em Journal of Machine Learning Research}, 1(3):1--48, 2010.
\newblock
  doi:{\color{blue}\href{https://doi.org/10.1016/j.websem.2011.06.001}{10.1016/j.websem.2011.06.001}}.

\bibitem{deep-graph-kernel}
Pinar Yanardag and SVN Vishwanathan.
\newblock Deep graph kernels.
\newblock In {\em Proceedings of the 21th ACM SIGKDD International Conference
  on Knowledge Discovery and Data Mining}, pages 1365--1374, 2015.
\newblock
  doi:{\color{blue}\href{https://doi.org/10.1145/2783258.2783417}{10.1145/2783258.2783417}}.

\bibitem{graphlet-kernel}
Nino Shervashidze, SVN Vishwanathan, Tobias Petri, Kurt Mehlhorn, and Karsten
  Borgwardt.
\newblock Efficient graphlet kernels for large graph comparison.
\newblock In {\em Artificial Intelligence and Statistics}, pages 488--495,
  2009.

\bibitem{node2vec}
Aditya Grover and Jure Leskovec.
\newblock node2vec: Scalable feature learning for networks.
\newblock In {\em Proceedings of the 22nd ACM SIGKDD international conference
  on Knowledge discovery and data mining}, pages 855--864, 2016.
\newblock
  doi:{\color{blue}\href{https://doi.org/10.1145/2939672.2939754}{10.1145/2939672.2939754}}.

\end{thebibliography}

\end{document}